\RequirePackage{fix-cm}
\documentclass[twocolumn]{svjour3}          
\smartqed  
\usepackage{graphicx}
\usepackage{amsmath}
\usepackage{cite}
\usepackage{amsmath,amssymb,amsfonts}
\usepackage{algorithmic}
\usepackage{graphicx}
\usepackage{textcomp}
\usepackage{hyperref}
\usepackage[super]{nth}
\usepackage{soul}
\usepackage{booktabs}
\usepackage{arydshln}
\usepackage{ulem}
\usepackage{multirow}

\newcommand{\orcid}[1]{\href{https://orcid.org/#1}{\includegraphics[scale=0.035]{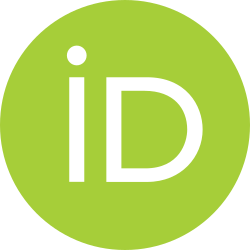}}}
\begin{document}

\title{HyPER-GAN: Hybrid Patch-Based Image-to-Image Translation for Real-Time Photorealism Enhancement in Game Engines
}


\author{Stefanos Pasios \orcid{0009-0002-9561-033X}         \and
        Nikos Nikolaidis \orcid{0000-0003-1515-7986}
}


\institute{Stefanos Pasios \at
              School of Informatics, Aristotle University of Thessaloniki (AUTH), Thessaloniki, 54124, Greece\\
              \email{pstefanos@csd.auth.gr}           
           \and
           Nikos Nikolaidis \at
              School of Informatics, Aristotle University of Thessaloniki (AUTH), Thessaloniki, 54124, Greece\\
              \email{nnik@csd.auth.gr}           
}

\date{Received: date / Accepted: date}

\maketitle

\begin{abstract}
Generative models are increasingly used in video game engines to enhance the photorealism of rendered images for visual synthetic data generation and simulation applications. However, they often introduce artifacts that alter the content of the original rendered scenes and require high computational resources, which limit their utilization for the photorealism enhancement of training and evaluation data, as well as their integration in the rendering pipelines of game engines. In this paper, we propose Hybrid Patch Enhanced Realism Generative Adversarial Network (HyPER-GAN), a hybrid image-to-image translation framework that is based on a lightweight U–Net–style generator capable of performing real-time inference. The framework is trained using paired rendered and photorealism-enhanced images, complemented by a novel hybrid training strategy that incorporates matched patches from unpaired real-world images to improve content preservation and further enhance the visual realism that can be achieved by the lightweight generator. Experimental results demonstrate that HyPER-GAN achieves a 6× increase in frames per second at 1080p in comparison with state-of-the-art lightweight paired image-to-image translation methods, while also increasing, in both within- and cross-engine evaluations, the photorealism of the rendered images without significantly compromising semantic consistency. Moreover, it is illustrated that HyPER-GAN maintains temporal consistency and that the proposed hybrid training strategy improves content preservation and visual realism in within-engine and increases the robustness in cross-engine evaluations compared to training the framework solely with paired rendered and photorealism-enhanced images. Code and pretrained models are publicly available at: \url{https://github.com/stefanos50/HyPER-GAN}

\end{abstract}

\section{Introduction}

Video game engines are widely used for deep learning research in order to train and evaluate Computer Vision (CV) algorithms in scenarios that are expensive, unsafe, or impractical to simulate in the real world \cite{pscs-i, NITA2026106013, seed4d}. While the photorealism of game engines has been improved by State-of-The-Art (SoTA) rendering technologies, such as Lumen of Unreal Engine 5 (UE5), there is still a significant difference between the rendered and real-world images, known as the simulation-to-reality (sim2real) appearance gap \cite{carla2real, sim2real_lane, pscs-i}, which often limits the real-world generalization performance of deep learning-based CV algorithms trained exclusively on rendered images.

Image-to-Image (Im2Im) translation \cite{cyclegan, pix2pixhd} has emerged as the primary approach for reducing the sim2real appearance gap, as it can enhance the photorealism of rendered images toward real-world characteristics while being less prone to visual artifacts (e.g., hallucinations) and significantly more computationally efficient than alternative generative approaches, such as diffusion models \cite{regen, aithal2024understandinghallucinationsdiffusionmodels}. Im2Im translation methods are categorized into unpaired \cite{fastcut, DCLGAN, Theiss2022UnpairedIT} and paired \cite{pix2pixhd, regen} approaches, depending on whether pixel-aligned source–target image pairs depicting the same content are available during training. For photorealism enhancement, unpaired Im2Im translation has been the dominant approach due to the practical difficulty of acquiring rendered–real-world image pairs with pixel-level correspondences, with methods such as DCLGAN \cite{DCLGAN} illustrating promising results for photorealism enhancement \cite{sim2real_lane}. More recent works proposed robust unpaired Im2Im translation approaches that further improved both visual fidelity and semantic consistency by utilizing additional inputs produced during image rendering, commonly referred to as G-Buffers (e.g., depth and surface normals) \cite{epe, 10458434, carla2real} that provide material, geometric, and semantic constraints and thus reduce the probability of generating artifacts (e.g., adding vegetation in the sky \cite{epe}). Since these G-Buffers require complex architectures that process them, these approaches are computationally expensive and slow at inference, running at 10 Frames Per Second (FPS) or below.

Due to the lack of paired rendered and real-world images, paired Im2Im translation focused on photorealism enhancement by synthesizing real-world images from semantic segmentation ground truth annotations generated in virtual environments (e.g., video games) \cite{10186767}. Considering that image synthesis is sensitive to the content distribution seen during training, this approach illustrated lower visual realism compared to the robust unpaired Im2Im translation methods \cite{epe}. To overcome the computational limitations of robust unpaired Im2Im translation methods that process additional information (i.e., G-Buffers) through complex deep learning architectures, the most recent paired Im2Im translation approach, REGEN \cite{regen}, reformulated the paired Im2Im translation strategy for photorealism enhancement. The reformulation involved the use of outputs from a robust unpaired Im2Im model \cite{epe} as proxy pairs for the real-world domain to regenerate the results with improved inference speed and without requiring the additional G-Buffers. However, the FPS achieved remained below real-time performance (30 FPS) at high resolutions (e.g., 1080p), and the framework could not reach or ideally surpass the semantic robustness of the initial robust unpaired Im2Im model, particularly due to the lack of an additional mechanism for avoiding learning the failure cases (artifacts) that can be potentially produced during the robust unpaired Im2Im translation phase \cite{Theiss2022UnpairedIT}. As a result, the photorealism-enhanced images may deviate from the original content of the rendered images, thereby limiting their use in CV algorithms, such as semantic segmentation \cite{Zhao2026} and object detection \cite{tian2025yolov12}, where maintaining semantic consistency (with the ground truth annotations) is of utmost importance.

\begin{table*}[h!]
\centering
\begin{tabular}{l c c c c c}
\hline
\textbf{Name} & \textbf{Year} & \textbf{GPU} & \textbf{Resolution} & \textbf{FPS} & \textbf{Training} \\
\hline
EPE \cite{epe} & 2022 & RTX 3090 & $957 \times 526$ & $2$ & Unpaired \\
EST-GAN \cite{9893673} & 2022 & - & - & - & Paired \\
NST \cite{10458434} & 2025 & - & 1080p & $10$ & Unpaired \\
CARLA2Real \cite{carla2real} & 2025 & RTX 4090 & $960 \times 540$ & $>$2 & Unpaired \\
REGEN \cite{regen} & 2026 & RTX 4090 & 1080p & $11$ & Paired \\
\hdashline
HyPER-GAN (proposed) & 2026 & RTX 4090 & 1080p & $78.90$ & Hybrid \\
\hline
\end{tabular}
\caption{Summary of photorealism enhancement methods.}
\label{tab:photorealism_methods}
\end{table*}

In this work, we build upon the paired Im2Im translation concept introduced by REGEN and propose \textit{Hybrid Patch Enhanced Realism Generative Adversarial Network (HyPER-GAN)}. In detail, HyPER-GAN employs a lightweight U-Net–style generator \cite{unet} that enables real-time inference on high-resolution rendered images. The generator is complemented by a new hybrid training strategy that incorporates both the photorealism-enhanced pairs and matched patches from the unpaired real-world dataset utilized during the robust unpaired Im2Im translation. This improves content preservation by avoiding the learning of artifacts produced by the robust unpaired Im2Im translation model, while further enhancing the visual realism of the lightweight U-Net–style generator, which has limited learning capacity. The experimental results demonstrate that HyPER-GAN achieves a $6x$ increase in FPS at 1080p without any model compression compared to SoTA lightweight Im2Im translation methods. In addition, it enhances the photorealism with improved semantic consistency in both within- and cross-engine evaluations that involve metrics that align with human perception and judgment. Moreover, it is demonstrated that HyPER-GAN can maintain an adequate level of temporal consistency. Finally, it is shown that the hybrid approach indeed improves the semantic consistency and visual realism in within-engine evaluation, while in parallel improving the robustness (better balance between semantic consistency and visual realism) in challenging unseen environments (cross-engine evaluation), compared to training with a variation of HyPER-GAN, HyPER-GAN Enhanced Only (HyPER-GAN-EO), which employs only paired rendered and photorealism-enhanced images. 

Our contributions are summarized as follows:
\begin{enumerate}
    \item We propose HyPER-GAN, an easy-to-integrate, lightweight Im2Im translation framework for real-time photorealism enhancement of rendered images.
    \item We introduce a hybrid training strategy that combines paired rendered–photorealism-enhanced image supervision with unpaired matched real-world image patches.
    \item We demonstrate through within- and cross-engine experiments that HyPER-GAN can enhance the photorealism of rendered images with a 6x improvement in FPS and better semantic consistency compared to SoTA lightweight paired Im2Im translation methods, while maintaining temporal consistency.
    \item We validate that the proposed hybrid training approach leads to superior semantic consistency, visual realism, and robustness in unseen environments, compared to training solely with rendered and photorealism-enhanced pairs.
\end{enumerate}

\section{Related Work}

Games such as Grand Theft Auto V (GTA-V) have become a popular source of rendered images for the training and evaluation of CV-based algorithms, motivating the development of appearance translation techniques \cite{epe, Theiss2022UnpairedIT, 9710598} that adapt these images towards the visual characteristics of real-world images. To this end, generative models \cite{DCLGAN, pix2pixhd, fastcut, cyclegan} have been extensively employed to enhance the photorealism of images generated from game engines and reduce the sim2real appearance gap across a wide range of application domains, such as autonomous driving \cite{carla2real,sim2real_lane}, robotics \cite{coholich2026sim2realimagetranslationenables}, crowd analysis \cite{pscs-i, wang2019learning}, and games \cite{regen}.

A major advancement in photorealism enhancement was introduced by Richter et al. in \cite{epe}, where the authors proposed Enhancing Photorealism Enhancement (EPE), an unpaired Im2Im translation method that utilizes additional G-Buffers (e.g., depth and albedo) generated by the game engines to improve visual realism, semantic consistency, and temporal stability. EPE achieves approximately 2 FPS at a resolution of $957\times526$ using an RTX 3090 GPU. Subsequently, building on the concept of EPE, Mittermueller et al. \cite{9893673} proposed EST-GAN, which extends CycleGAN (unpaired Im2Im) \cite{cyclegan} and Pix2PixHD \cite{pix2pixhd} (paired Im2Im) to utilize G-Buffers for translating Unity-rendered images toward the visual appearance of Red Dead Redemption 2. This is done by combining the output of CycleGAN with Pix2PixHD. The real-time integration of EST-GAN was left as future work. Ioannou and Maddock \cite{10458434} introduced an unpaired Neural Style Transfer (NST) approach that incorporates motion information alongside additional rendering information (i.e., G-Buffers) to improve temporal consistency, achieving approximately 10 FPS at 1080p on unspecified hardware.

Pasios and Nikolaidis \cite{carla2real} proposed CARLA2Real, a tool that employes a modified lightweight version of the EPE architecture designed for integration within the CARLA simulator \cite{carla}. Specifically, CARLA2Real aligns the rendered images of CARLA with real-world datasets such as Cityscapes (CS) \cite{cityscapes} and Mapillary Vistas (MV) \cite{vistas} in an unpaired Im2Im translation manner, achieving slightly above 2 FPS at $960\times540$ resolution on an RTX 4090 GPU without further optimization (e.g., TensorRT). Finally, more recently, the same authors introduced REGEN \cite{regen}, a two-stage framework that utilizes the output of a robust unpaired Im2Im translation method (i.e., EPE) for training a more lightweight paired Im2Im translation approach that is capable of performing (near) real-time inference. REGEN achieves 11 FPS at 1080p using an RTX 4090 GPU.

Despite these advances, existing methods, as summarized in Table \ref{tab:photorealism_methods}, remain considerably below the typical real-time target of 30 FPS at 1080p. Furthermore, they either rely on computationally demanding unpaired Im2Im translation (EPE, NST, and CARLA2Real) or require an additional unpaired Im2Im translation model to generate paired training data (EST-GAN and REGEN), which may introduce artifacts into the training process of the paired Im2Im translation method. To address these limitations, we propose HyPER-GAN, which is based on a lightweight U-Net-style network that enables higher inference speeds and is coupled with a hybrid training strategy that leverages both paired proxy photorealism-enhanced images and unpaired real-world images, mitigating artifact propagation from the unpaired Im2Im translation models and therefore improving the 
lightweight U-Net-style network in terms of visual realism and robustness on unseen environments.

\begin{figure*}[htbp]
    \centering
    \includegraphics[width=1\textwidth]{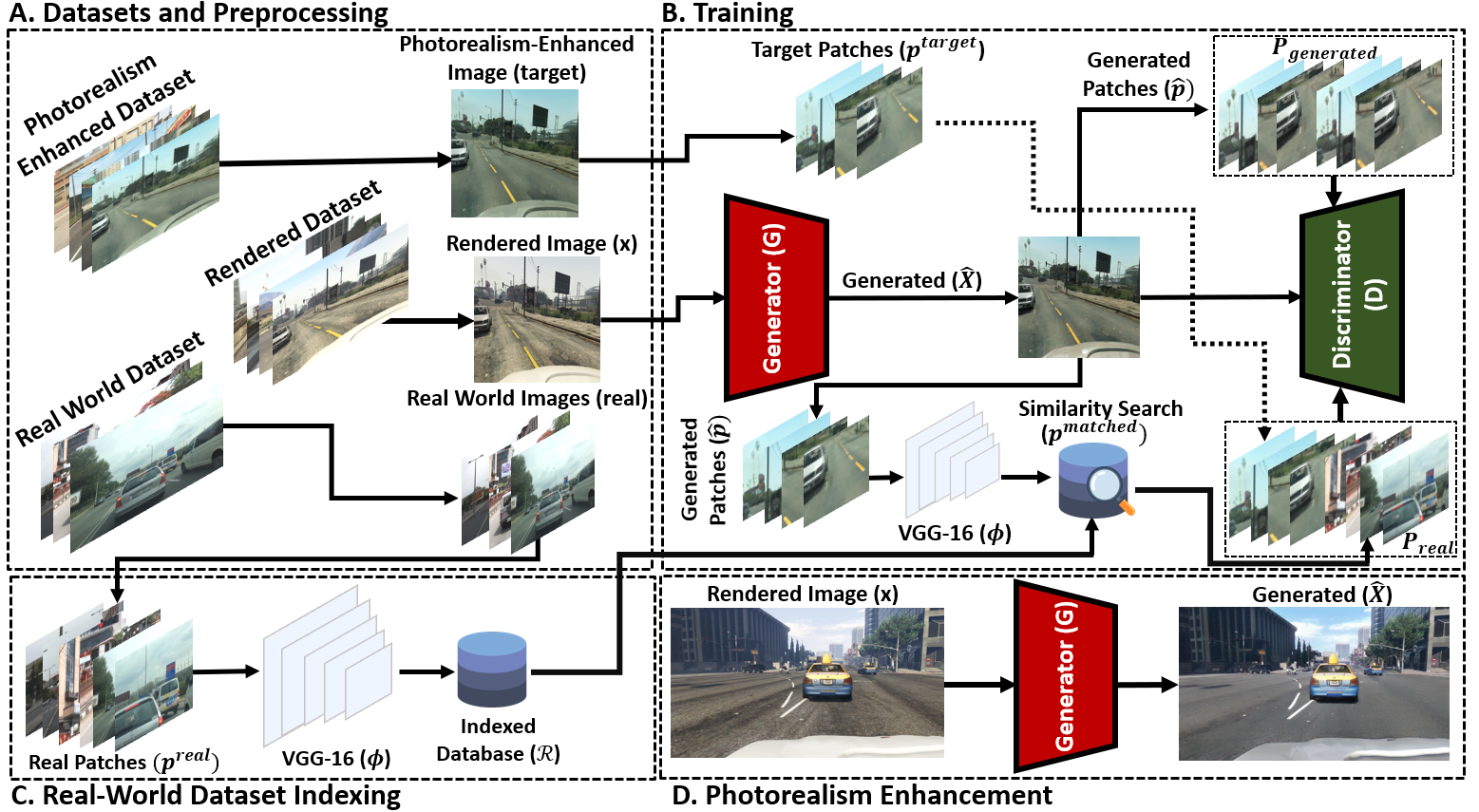}
    \caption{Overview of the HyPER-GAN framework, which includes four phases: a) datasets and preprocessing, b) real-world dataset indexing, c) training, and d) photorealism enhancement.}
    \label{fig:flowchart}
\end{figure*}

\section{HyPER-GAN}

The HyPER-GAN framework includes four phases, as illustrated in Figure \ref{fig:flowchart}, namely the datasets and preprocessing, the real-world dataset indexing, the training, and the photorealism enhancement phases. These are detailed in the following subsections.

\subsection{Datasets and Preprocessing}

HyPER-GAN requires three types of datasets: (i) a rendered dataset containing rendered images ($x$) from simulators, e.g., CARLA \cite{carla} or video games such as Grand Theft Auto V (GTA-V); (ii) a photorealism-enhanced dataset including photorealism-enhanced image pairs ($target$) of the rendered dataset generated by a robust unpaired Im2Im translation model, such as EPE \cite{epe}, using real-world images as targets; and (iii) the real-world dataset that incorporates the real-world images ($real$) used as the target of realism during the robust unpaired Im2Im translation. All the images included in the datasets are preprocessed using the same pipeline. In detail, images are first resized to a fixed resolution of 512x512, which is a standard resolution employed by paired Im2Im translation methods such as Pix2PixHD \cite{pix2pixhd}. Subsequently, each image is converted to a tensor and normalized using a mean and standard deviation of [0.5, 0.5, 0.5].

\subsection{Real-World Dataset Indexing}

To effectively incorporate real-world images in HyPER-GAN and considering the issues in unpaired Im2Im translation, where distributional object differences can result in visual artifacts since the discriminator will learn to distinguish real and generated images by these distribution differences \cite{epe}, we follow a patch matching approach proposed in a previous photorealism enhancement unpaired Im2Im method (i.e., EPE). To this end, the Facebook AI Similarity Search (FAISS) \cite{faiss} library is employed for fast nearest-neighbor search in high-dimensional spaces. In detail, we extract four $196 \times 196$ non-overlapping patches (after resizing to $512 \times 512$) from each $real$ image of the real-world dataset (these patches are denoted as $p^{real}$ in Figure \ref{fig:flowchart}) and subsequently compute feature embeddings $\phi(p^{real})$ for each patch using a pre-trained VGG-16 \cite{vgg} backbone (the third convolutional layer in VGG-16 block 4), which are stored in the FAISS indexed database ($\mathcal{R}$) using $L_2$ distance.

\subsection{Training}

Training follows the standard Generative Adversarial Network (GAN) process, where a generator ($G$) aims to generate photorealism-enhanced images, while a discriminator ($D$) attempts to classify whether an input image originates from the generator or the target domain dataset.

\paragraph{Generator and Discriminator:} 
$G$ is a lightweight U-Net-style network (an architecture widely employed for GANs \cite{BATZIOU2026105889, Yao2025}), and $D$ is a PatchGAN-style network operating on patches. In detail, $G$ consists of an encoder (three downsampling stages) and a decoder (three upsampling stages). The encoder receives a rendered image $x \in \mathbb{R}^{3 \times H \times W}$ and progressively increases the feature dimensionality from 3 to 256 channels (64, 128, and 256) using strided convolutions (denoted as Conv in the following equations). Instance Normalization  (IN) and Rectified Linear Unit (ReLU) activation are applied after all encoder layers, except for IN after the first, as shown in Eq. \ref{eq:encoder_layers}. 

\begin{equation}
\label{eq:encoder_layers}
\begin{aligned}
e_1 &= \mathrm{ReLU}(\mathrm{Conv}_{4\times4}^{s=2,p=1}(x)) \\
e_2 &= \mathrm{ReLU}(\mathrm{IN}(\mathrm{Conv}_{4\times4}^{s=2,p=1}(e_1))) \\
e_3 &= \mathrm{ReLU}(\mathrm{IN}(\mathrm{Conv}_{4\times4}^{s=2,p=1}(e_2)))
\end{aligned}
\end{equation}

A bottleneck composed of four residual blocks enables deeper feature extraction while preserving spatial information through identity skip connections. In detail, each residual block contains two convolutional layers with IN, using a ReLU activation after the first convolution and an identity skip connection as detailed in Eq. \ref{eq:bottleneck}. 
 
\begin{equation}
\label{eq:bottleneck}
\begin{aligned}
m &= \mathrm{ResBlock}^{\times4}(e_3) \\
\mathrm{ResBlock}(z) &= z + \mathrm{IN}(\mathrm{Conv}_{3\times3}^{1,1}(
\mathrm{ReLU}(\mathrm{IN}(\mathrm{Conv}_{3\times3}^{1,1}(z)))))
\end{aligned}
\end{equation}
 
 The decoder mirrors the encoder using transposed convolutions and skip connections via feature concatenation from corresponding encoder stages. IN and ReLU activation are applied in intermediate decoder layers, and a final Tanh activation produces the normalized generated image $\hat{X} \in \mathbb{R}^{3 \times H \times W}$ as illustrated in Eq. \ref{eq:decoder} 

\begin{equation}
\label{eq:decoder}
\begin{aligned}
d_3 &= \mathrm{ReLU}(\mathrm{IN}(\mathrm{ConvTranspose}_{4\times4}^{s=2,p=1}(m))) \\
d_2 &= \mathrm{ReLU}(\mathrm{IN}(\mathrm{ConvTranspose}_{4\times4}^{s=2,p=1}([d_3 \Vert e_2]))) \\
G(x) = \hat{X} &= \mathrm{Tanh}(\mathrm{ConvTranspose}_{4\times4}^{s=2,p=1}([d_2 \Vert e_1]))
\end{aligned}
\end{equation}

where $[\cdot \Vert \cdot]$ denotes channel-wise concatenation.
 
The discriminator $D$ follows a PatchGAN-style architecture that evaluates realism at the patch level. Given a patch $p \in \mathbb{R}^{3 \times H \times W}$, the $D$ consists of a sequence of strided convolutional layers with increasing feature dimensionality (64, 128, and 256), progressively reducing spatial resolution while capturing local texture statistics, each followed by Leaky ReLU activation. IN is applied to all layers except the first. A final 1×1 convolution outputs a single-channel feature map representing the realism scores of local image patches as presented in Eq. \ref{eq:discriminator}.
 
\begin{equation}
\label{eq:discriminator}
\begin{aligned}
h_1 &= \mathrm{LeakyReLU}_{0.2}(\mathrm{Conv}_{4\times4}^{s=2,p=1}(p)) \\
h_2 &= \mathrm{LeakyReLU}_{0.2}(\mathrm{IN}(\mathrm{Conv}_{4\times4}^{s=2,p=1}(h_1))) \\
h_3 &= \mathrm{LeakyReLU}_{0.2}(\mathrm{IN}(\mathrm{Conv}_{4\times4}^{s=2,p=1}(h_2))) \\
D(p) &= \mathrm{Conv}_{4\times4}^{s=1,p=1}(h_3)
\end{aligned}
\end{equation}

\paragraph{Similarity Search} 
For each generated image $\hat{X} = G(x)$, a set of four $196\times196$ non-overlapping patches $\hat{p}$ is extracted. Subsequently, for every generated patch $\hat{p}$, the spatially corresponding patch $p^{target}$ from the photorealism-enhanced image ($target$) pair is retrieved. To avoid learning artifacts from $target$, the FAISS indexed database ($\mathcal{R}$) is also employed to find (match) the nearest neighbor $p^{real} \in \mathcal{R}$ of $\hat{p}$ in the VGG-16 feature space:
    \begin{equation}
    p^{mached} = \arg\min_{p^{real} \in \mathcal{R}} \| \phi(\hat{p}) - \phi(p^{real}) \|_2^2
    \end{equation}

\begin{figure}
	\centering
	\includegraphics[width=1\columnwidth]{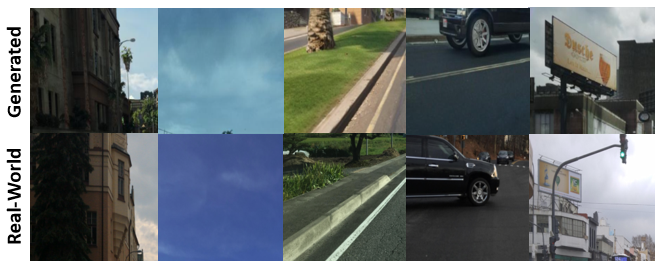}
	\caption{Examples of matched patches between the generated (top) and the real-world (bottom).}
	\label{fig:matched}
\end{figure}

Examples of such matched patches are illustrated in Figure \ref{fig:matched}, where it is evident that semantically similar content (e.g., sky, building, or advertisement board) is matched between the generated and the real-world images. Following the extraction of matched patches, we form two sets of batches that contain eight patches (since four patches are extracted from each image): a $generated$ set $\mathcal{P}_{generated} = [\hat{p}, \hat{p}]$ and a $real$ set $\mathcal{P}_{real} = [p^{target}, p^{mached}]$. By processing these sets of batches, $D$ is forced to distinguish $G$'s output from both the enhanced ($target$) and the real-world ($real$) domains. Particularly, this additional real-world supervision discourages $G$ from learning artifacts introduced by the robust unpaired Im2Im translation model, as it must fool $D$ not only on the photorealism-enhanced target domain but also on the real-world one, where such artifacts are absent but as well as to improve the overall visual realism. For comparison, we also consider a simplified variation, HyPER-GAN-EO, in which $G$ is trained exclusively on the paired rendered–photorealism-enhanced images ($\mathcal{P}_{generated} = [\hat{p}]$ and $\mathcal{P}_{real} = [p^{target}]$).

\paragraph{Loss Functions:} 
$G$ is trained using a combination of adversarial and reconstruction losses. The adversarial component follows the Least-Squares GAN (LSGAN) formulation~\cite{8237566}, which replaces the binary cross-entropy objective with a least-squares loss to stabilize training and improve gradient quality. $D$ is trained to assign a value of $1$ to real patches $\mathcal{P}_{real}$ and $0$ to generated patches $\mathcal{P}_{generated}$. Conversely, $G$ is trained to produce patches that are classified as real by $D$, i.e., to push $D(\mathcal{P}_{generated})$ toward $1$. In addition, an $L_1$ reconstruction loss between the generated image $\hat{X}$ and the photorealism-enhanced image $target$ is used to preserve structural and semantic consistency. The overall objectives are:

\begin{equation}
\begin{aligned}
\mathcal{L}_D &=
\mathbb{E}_{q \sim \mathcal{P}_{real}} \big[ (D(q)-1)^2 \big]
+ \mathbb{E}_{q \sim \mathcal{P}_{generated}} \big[ D(q)^2 \big] \\
\mathcal{L}_G &=
\mathbb{E}_{q \sim \mathcal{P}_{generated}} \big[ (D(q)-1)^2 \big]
+ \lambda \, \| \hat{X} - target \|_1
\end{aligned}
\end{equation}

where $\lambda=10$ is a weighting factor for $L_1$ distance.

\subsection{Photorealism Enhancement}

\begin{figure}
	\centering
	\includegraphics[width=1\columnwidth]{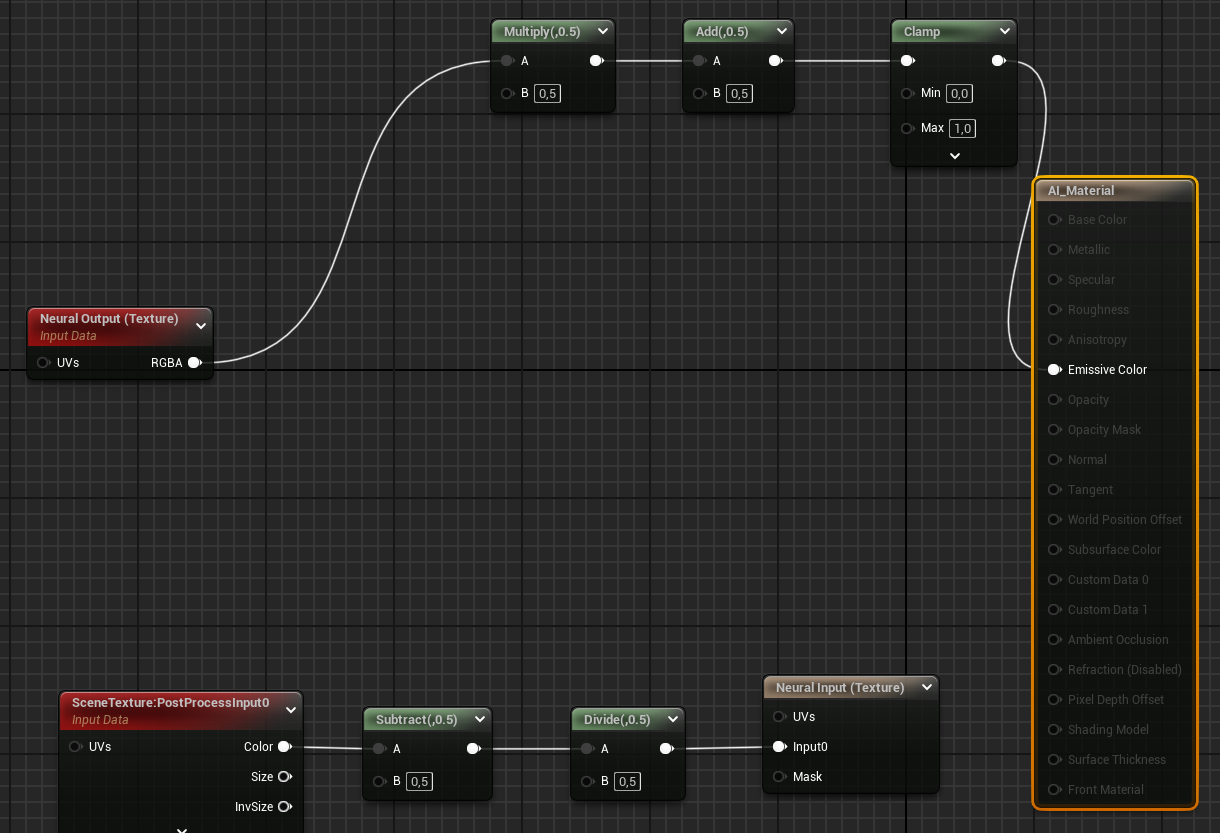}
	\caption{Setup of the HyPER-GAN post-processing material inside UE5.}
	\label{fig:setup}
\end{figure}

During the photorealism enhancement phase, HyPER-GAN operates as a standalone feed-forward network. In detail, the FAISS index and $D$ are discarded, and given a rendered input image $x$, the $G$ directly produces the photorealism-enhanced output $\hat{X}$. HyPER-GAN does not require any additional input, such as G-Buffers or semantic segmentation ground truth annotations that are typically expected by robust unpaired Im2Im translation networks \cite{epe, 10458434}, and therefore can be easily integrated into the rendering pipelines of game engines as a post-processing filter. In particular, for the UE5 game engine since version 5.4 and above, the engine natively supports running neural rendering models using the Open Neural Network Exchange (ONNX) format by enabling the \textbf{Neural Rendering} plugin\footnote{\url{https://dev.epicgames.com/documentation/unreal-engine/neural-post-processing-in-unreal-engine}}. Therefore, HyPER-GAN, considering that it also does not require additional inputs such as G-Buffers, can be easily integrated. In detail, in Figure \ref{fig:setup}, the exact preprocessing (input) and post-processing (output) steps that should be applied for HyPER-GAN inside the post-processing material of UE5 are provided. 

\section{Experiments and Results}

In this section, first, the datasets and evaluation metrics are detailed, and subsequently, the experimental setup used to illustrate the contributions of HyPER-GAN is described. Next, the implementation details are outlined, and, finally, the results are presented and discussed. 

\subsection{Datasets}

To conduct the experiments, both datasets rendered within a game engine, as well as datasets captured in the real-world, are required. The characteristics of the employed datasets are detailed below.

\paragraph{Rendered Datasets:} Playing For Data (PFD)\footnote{\url{https://download.visinf.tu-darmstadt.de/data/from_games/}} \cite{pfd} is a dataset that was rendered using the GTA-V video game, which is based on the Rockstar Advanced Game Engine
(RAGE) game engine. The dataset contains a total of $25,000$ rendered images that are accompanied by ground truth semantic segmentation annotations that follow the    CS \cite{cityscapes} dataset annotation scheme (35 classes). Virtual KITTI 2 (VKITTI2)\footnote{\url{https://europe.naverlabs.com/proxy-virtual-worlds-vkitti-2/}} \cite{cabon2020virtualkitti2} is a dataset that clones five different scenes that are included in the real-world KITTI \cite{kitti}  (e.g., a vehicle in an image of the real-world KITTI dataset will have roughly the same position, orientation, and properties also in the corresponding VKITTI2 rendered image) dataset using the Unity game engine. The dataset includes a total of $2,126$ unique clone images that are accompanied by various annotations, including bounding boxes and semantic segmentation annotations (15 classes).  The OffRoadSynth\footnote{\url{https://www.kaggle.com/datasets/konrmal94/synthetic-offroad-image-dataset}} \cite{OffRoadSynth}  dataset is also captured within the Unity game engine, containing $6,260$ images with semantic segmentation annotations (5 classes). Unlike VKITTI2, which depicts urban scenes, OffRoadSynth is captured in an off-road environment. CrowdFlow\footnote{\url{https://github.com/tsenst/CrowdFlow}} is a dataset that was generated using the Unreal Engine 4 (UE4) game engine and was designed for evaluating the performance of optical flow estimation methods. The dataset incorporates $10$ videos ($5$ from static and $5$ from dynamic drone cameras) that span a total of $3,600$ images. Each image is accompanied by ground truth annotations for optical flow fields, person trajectories, and dense pixel trajectories.

\paragraph{Real-World Datasets:} CS\footnote{\url{https://www.cityscapes-dataset.com/}} \cite{cityscapes} is a real-world dataset captured in $50$ different cities across Germany and contains  semantic segmentation annotation with $35$ classes. CS is characterized by the dark and greyish color tones that resulted from the employed camera, as well as the Mercedes auto manufacturer logo, which is depicted in all the $5,000$ images of the dataset and has been proven to challenge Im2Im translation methods \cite{epe}. MV\footnote{\url{https://www.mapillary.com/dataset/vistas}} \cite{vistas} is another real-world dataset that was captured and annotated for semantic segmentation tasks with a total of $25,000$ images and $124$ classes. Unlike  CS, which is limited to Germany, MV includes images from locations across $6$ continents.

\subsection{Metrics}

A robust assessment of visual realism demands human-centric evaluation; we have therefore adopted metrics with proven perceptual alignment. To this end, the CLIP Maximum Mean Discrepancy (CMMD) \cite{cmmd} metric, which employs the rich feature space of CLIP \cite{NEURIPS2022_904aac1c}, was selected for visual realism quantification. CMMD measures the distributional distance between image sets in the CLIP feature space, where lower values indicate closer feature alignment to real images and, as a result, better visual realism. Through user studies, CMMD has been proven to align well with human visual perception. In addition, Learned Perceptual Image Patch Similarity (LPIPS) \cite{lpips} was employed as a complementary perceptual metric, measuring the degree of change of the visual content,  using the initial rendered image as reference and comparing image features, focusing on differences in texture, structure, and overall appearance as perceived by humans. In a photorealism enhancement setup, lower LPIPS is better. Furthermore, to investigate objects resulting from potential hallucinations, Intersection over Union (IoU) was employed to measure how well the predictions of a semantic segmentation model on the photorealism-enhanced images align with the ground truth annotations of the rendered images (higher is better). Finally, the end-point error (lower is better), which quantifies the average Euclidean distance between the estimated motion vectors and the corresponding ground truth motion vectors, was used to measure temporal consistency.

For photorealism enhancement, LPIPS and IoU are the most important metrics since distortions, artifacts, and hallucinations can degrade the performance of CV algorithms trained on the photorealism-enhanced images due to the misalignment between the visual content and the ground truth annotations. Therefore, better visual realism (i.e., lower CMMD) should also be accompanied by higher semantic consistency (i.e., lower LPIPS and higher IoU).

\subsection{Experimental Setup}

The experimental setup detailed in this section includes real-time benchmarking, within-engine and cross-engine evaluations, as well as a temporal consistency experiment.

\paragraph{Real-Time Benchmarking:} The real-time performance capabilities of the proposed HyPER-GAN and other baselines (i.e., FastCUT and REGEN) were first assessed in terms of the inference latency and FPS, reported as mean $\pm$ standard deviation over 100 images, as well as the VRAM utilization on multiple resolutions (for real-time benchmarking within UE5, see Appendix \ref{app: benchmarkingUE5}). The HyPER-GAN-EO variation was not considered since it is based on the exact same $G$ as HyPER-GAN (only the $G$ is used during inference, as illustrated in Figure \ref{fig:flowchart}). In detail, benchmarking was performed on a system equipped with an Intel i7-14700F CPU, 32GB DDR4 RAM, and an NVIDIA RTX 4090 GPU with 24GB of VRAM at 720p, 1080p, 1440p, and 4K resolutions. Furthermore,  a lower specification system with an Intel i7 13700KF CPU, an NVIDIA RTX 4070 Super GPU with 12GB of VRAM, and 32GB of DDR4 system memory was also used to perform experiments at 720p and 1080p resolutions. The benchmarking was conducted without using any model compression or optimization (e.g., TensorRT).

\paragraph{Within-engine:} This experiment involved the utilization of a dataset that was extracted from a virtual environment to perform a within-engine evaluation. To this end, the $25,000$ images of the Playing for Data (PFD) \cite{pfd} dataset were employed. In addition, we utilized the photorealism-enhanced counterparts for a subset of $19.252$ of these images, translated by EPE towards the characteristics of the real-world datasets CS and MV. These are provided in \cite{epe} and were used as the photorealism-enhanced pairs generated from a robust unpaired Im2Im translation model. In detail, in order to transform these datasets into a compatible dataset for paired Im2Im translation, the  $19,252$ images from the  PFD dataset and their photorealism-enhanced counterparts were resized to a resolution of $512 \times 512$  to match the resolution required during training by paired Im2Im translation methods \cite{pix2pixhd}, including HyPER-GAN. Subsequently, the $19,252$ original-photorealism-enhanced pairs were split using the official split provided by the authors. This resulted in $9,549$ training, $4,987$ validation, and $4,716$ test images. Considering that the photorealism-enhanced pairs target the CS and MV real-world datasets, the images of these datasets were employed as the real-world images required by the proposed framework. Specifically, we utilized the $5,000$ images provided in the CS and the $25,000$ included in MV. 

Having constructed the required datasets, the HyPER-GAN and the HyPER-GAN-EO variation were trained to also learn to perform photorealism enhancement from the rendered domain (GTA-V) towards the photorealism-enhanced outputs of EPE (CS and MV). The same training procedure was also performed for two baseline lightweight paired Im2Im translation methods, namely FastCUT \cite{fastcut} and REGEN \cite{regen}. Subsequently, the test set of the dataset (i.e., $4,716$ images) was photorealism-enhanced with each of the trained models, and the visual realism was evaluated using the target real-world datasets and the CMMD metric. As an example, images photorealism-enhanced towards CS were evaluated against CS. In addition, the LPIPS metric was calculated using the initial  PFD (GTA-V) images and the photorealism-enhanced ones produced by FastCUT, REGEN, HyPER-GAN-EO, and HyPER-GAN.
To further investigate the semantic preservation capabilities of the proposed HyPER-GAN, the HyPER-GAN-EO variation, and the baseline methods, Mask2Former \cite{mask2former} semantic segmentation models, pretrained on the real-world datasets (i.e., CS and MV) using the official weights\footnote{\url{https://huggingface.co/facebook/mask2former-swin-large-cityscapes-semantic}}\footnote{\url{https://huggingface.co/facebook/mask2former-swin-large-mapillary-vistas-semantic}}, were also applied to the photorealism-enhanced images, and their predictions were evaluated against the ground truth annotations using mIoU. A reduction in semantic segmentation accuracy compared to the one achieved with the initial images highlights that images generated by a certain method are subject to more visual artifacts and visual inconsistencies, such as hallucinations of objects. Since there are incompatibilities between the pretrained semantic segmentation models and the PFD dataset, we employed the typical 19 CS classes used for benchmarking semantic segmentation models \cite{cityscapes} and set the remaining object classes as background.

\paragraph{Cross-engine:} Cross-engine evaluation on rendered images of a different game engine was also conducted. Specifically, the VKITTI2 \cite{cabon2020virtualkitti2} and OffRoadSynth \cite{OffRoadSynth} datasets, which are based on the Unity game engine, were employed. In detail, to perform the cross-engine experiment, FastCUT, REGEN, HyPER-GAN-EO, and HyPERGAN models trained  (in the within-engine experiment) to translate PFD towards CS and MV  were applied on the VKITTI2 and OffRoadSynth images (without any resizing), and the CMMD and LPIPS metrics were calculated similarly to the within-engine experiment using the PFD dataset. Specifically, the $2,126$ clone images of the VKITTI2 were utilized, and for the OffRoadSynth, where the authors provide pre-defined train, validation, and test sets, the $1,234$ images included in the test set were employed. In addition, the Mask2Former models (CS and MV) were applied for the calculation of the mIoU. To enable compatibility between the pretrained Mask2Former models and the VKITTI2 classes, we merged the following pairs of classes: vegetation with tree, truck with van, and misc with unlabeled, with the latter not considered for evaluation since it doesn't exist in the Mask2Former pretrained CS and MV models. This resulted in a total of 11 object classes. For  OffRoadSynth, from the $5$ classes, the trees/vegetation, grass, and sky classes were preserved while the path and obstacles were merged as background (and were not considered in the evaluation), since they don't exist in the pretrained Mask2Former models.

\paragraph{Temporal Consistency: } We also investigated the temporal consistency of the HyPER-GAN framework since it is another important factor in real-time applications such as video games \cite{regen} and simulations \cite{Martin2026}. Along those lines, we performed the same experiment that was conducted for REGEN in \cite{regen}. In detail, the CrowdFlow \cite{crowd-flow} dataset, which includes ground truth optical flow annotations,  was employed. Subsequently, the HyPER-GAN models trained to translate PFD towards the CS and MV characteristics in the within-engine experiment were applied to the dataset in a cross-engine manner since CrowdFlow is based on the UE4 game engine. Optical flow predictions were calculated using a traditional optical flow estimation algorithm, namely, Farneback's algorithm \cite{Farneb}, on both the initial, temporally consistent, rendered images and the photorealism-enhanced ones (CS and MV), and the error compared to the ground truth optical flows was evaluated using the endpoint error metric. For evaluation, as in \cite{regen}, we utilized only the sequences of the CrowdFlow dataset, which include dynamic camera (drone) movement, since they are more prone to visual artifacts, and the results were split for the pixels that depict the background and the ones that show the crowd. The rationale behind this experiment is that the endpoint error on the rendered and the photorealism-enhanced images should remain similar. 

\begin{figure}
	\centering
	\includegraphics[width=1.00\columnwidth]{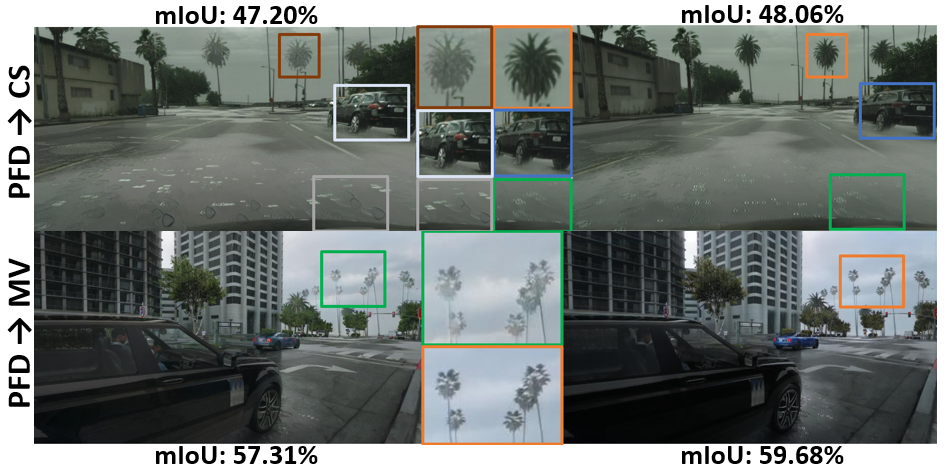}
	\caption{Translation results of PFD (GTA-V) towards real-world datasets (CS and MV) produced by EPE (left) and HyPER-GAN (right) on images from the test set. Mask2Former mIoU is also reported for the entire test sets.}
	\label{fig:failures}
\end{figure}

\begin{figure*}
	\centering
	\includegraphics[width=1\textwidth]{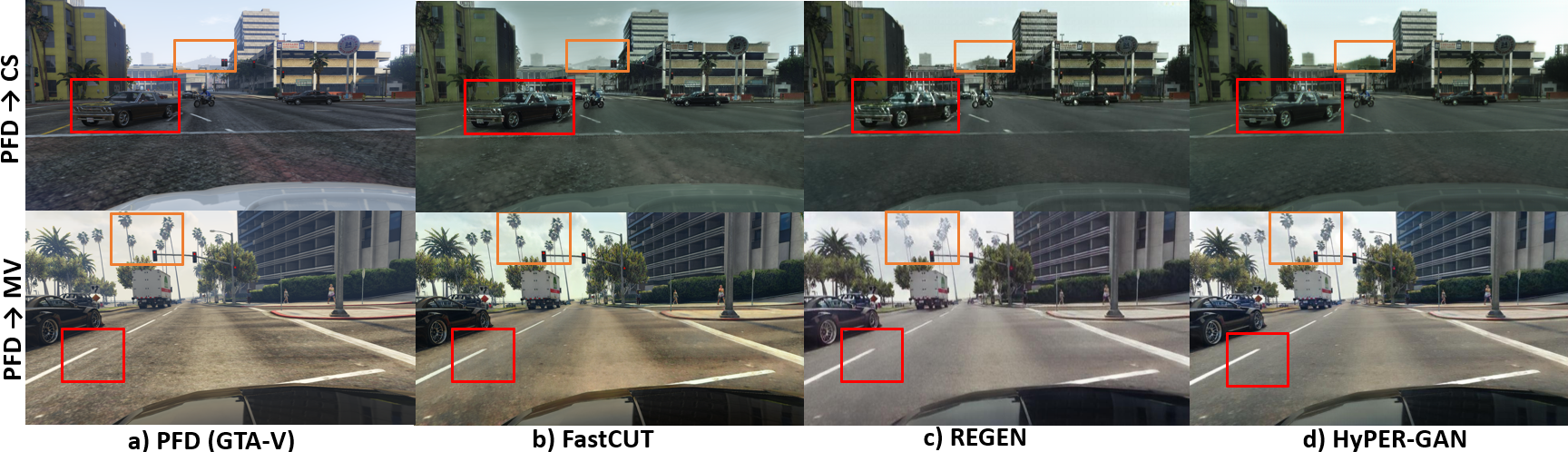}
	\caption{Translation results of a) PFD (GTA-V) towards the real-world datasets CS and MV produced by b) FastCUT, c) REGEN, and d) HyPER-GAN.}
	\label{fig:baselines_comp_img}
\end{figure*}

\subsection{Implementation Details}

The HyPER-GAN and the HyPER-GAN-EO variation were implemented in PyTorch and trained on a single NVIDIA RTX 4070 GPU with 12GB of memory. $G$ and $D$ were optimized using the Adam optimizer with a learning rate of $2\times10^{-4}$ and betas $(0.5, 0.999)$. We trained HyPER-GAN and HyPER-GAN-EO for 20 epochs with a batch size of 1. For REGEN, we used the official\footnote{\url{https://github.com/stefanos50/REGEN}} implementation and trained for 9 epochs, as in \cite{regen}. FastCUT was also employed with the official\footnote{\url{https://github.com/taesungp/contrastive-unpaired-translation}} implementation and was trained for 20 epochs.

\begin{table}[ht]
\centering
\resizebox{\columnwidth}{!}{%
\begin{tabular}{l c c c c}
\hline
Model & Resolution & $\downarrow$ Latency (ms) & $\uparrow$ FPS & $\downarrow$ VRAM (GB) \\
\hline
FastCUT & 720p & $235.718 \pm 10.925$ & $4.25 \pm 0.18$ & 2.6 \\
& 1080p & $531.625 \pm 1.246$ & $1.88 \pm 0.00$ & 5.1 \\
& 1440p & $959.781 \pm 17.422$ & $1.04 \pm 0.02$ & 8.1 \\
& 4K & - & - & Out of Memory \\
\hline
REGEN & 720p & $36.247 \pm 0.288$ & $27.59 \pm 0.21$ & 2.2 \\
& 1080p & $79.520 \pm 4.634$ & $12.61 \pm 0.63$ & 3.8 \\
& 1440p & $149.088 \pm 4.831$ & $6.71 \pm 0.21$ & 5.0 \\
& 4K & $353.814 \pm 11.913$ & $2.83 \pm 0.09$ & 10.3 \\
\hline
HyPER-GAN & 720p & $5.281 \pm 0.192$ & $189.58 \pm 6.40$ & 1.0 \\
& 1080p & $12.680 \pm 0.274$ & $78.90 \pm 1.58$ & 1.8 \\
& 1440p & $25.483 \pm 0.392$ & $39.25 \pm 0.57$ & 2.8 \\
& 4K & $63.051 \pm 0.538$ & $15.86 \pm 0.13$ & 5.6 \\
\hline
\end{tabular}%
}
\caption{Runtime performance (latency, FPS) and VRAM usage of FastCUT, REGEN, and HyPER-GAN across different resolutions on an NVIDIA RTX 4090 GPU. \label{tab:performance_comparison}}
\end{table}

\begin{table}[ht]
\centering
\resizebox{\columnwidth}{!}{%
\begin{tabular}{l c c c c}
\hline
Model & Resolution & $\downarrow$ Latency (ms) & $\uparrow$ FPS & $\downarrow$ VRAM (GB) \\
\hline
FastCUT & 720p & $128.210 \pm 0.467$ & $7.80 \pm 0.032$ & 2.3 \\
& 1080p & $297.942 \pm 3.516$ & $3.36 \pm 0.04$ & 3.8 \\
\hline
REGEN & 720p & $79.175 \pm 0.770$ & $12.63 \pm 0.12$ & 1.9 \\
& 1080p & $180.973 \pm 1.937$ & $5.53 \pm 0.06$ & 3.1 \\
\hline
HyPER-GAN & 720p & $12.347 \pm 0.279$ & $81.03 \pm 1.80$ & 0.8 \\
& 1080p & $29.642 \pm 0.175$ & $33.74 \pm 0.20$ & 1.5 \\
\hline
\end{tabular}%
}
\caption{Runtime performance (latency, FPS) and VRAM usage of FastCUT, REGEN, and HyPER-GAN across different resolutions on an NVIDIA RTX 4070 Super GPU. \label{tab:performance_comparison2}}
\end{table}

\subsection{Results and Discussion} \label{sec:disc}

In this subsection, we first present the results of FastCUT, REGEN, and HyPER-GAN in terms of their real-time capabilities. Subsequently, the semantic robustness and visual realism results in both within- and cross-engine evaluations of FastCUT, REGEN, HyPER-GAN-EO, and HyPER-GAN are reported and discussed. Finally, the findings regarding the temporal consistency of HyPER-GAN are showcased. 

\paragraph{Real-Time Benchmarking Results: } 
Table \ref{tab:performance_comparison} presents the inference latency and FPS, reported as mean $\pm$ standard deviation over 100 images, as well as the VRAM utilization on multiple resolutions (i.e., 720p, 1080p, 1440p, and 4K) for FastCUT, REGEN, and HyPER-GAN using the system with the RTX 4090 GPU. As demonstrated, the proposed framework achieves real-time performance, reaching the target of 30 FPS at 1440p and a (near) real-time performance of 15 FPS at 4K. On the other hand, FastCUT and REGEN are incapable of achieving real-time FPS at any resolution (i.e., 720p, 1080p, 1440p, and 4K) while they require roughly double the VRAM compared to HyPER-GAN. Precisely, at 1080p, which is a standard resolution, HyPER-GAN provides a 6x increase in FPS ($12.61$ to $78.90$ FPS) and a $2x$ reduction in VRAM ($3.8$ to $1.8$ GB) compared to REGEN. For the lower specification system with the RTX 4070 Super GPU, the results are illustrated in Table \ref{tab:performance_comparison2}. It is again evident that HyPER-GAN, compared to FastCUT and REGEN, is the only method that can achieve 30 FPS at  1080p on a mid-range GPU such as the RTX 4070 Super.

\paragraph{Within-engine Results: } For the within-engine evaluation on the PFD (GTA-V), the results are shown in Table \ref{tab:exp1}. As illustrated, HyPER-GAN achieves the best or the second-best score across all the evaluation metrics. In detail, compared to FastCUT, HyPER-GAN achieves better visual realism (CMMD) as well as content preservation (LPIPS and mIoU) for both CS and MV. Moreover, in comparison with REGEN, it achieves a higher degree of semantic consistency (lower LPIPS and higher mIoU) while effectively enhancing the photorealism by reducing the CMMD metric of the initial rendered PDF images from $5.274$ to $3.080$ for CS and from $4.296$ to $2.827$ for MV. HyPER-GAN also outperforms the HyPER-GAN-EO variation, which is trained solely using rendered and photorealism-enhanced images across most of the metrics for the CS and all of the metrics in MV photorealism-enhanced images, which highlights the contribution of the introduced hybrid training strategy (improved semantic consistency and visual realism). The increased visual realism compared to the rendered PFD (GTA-V) images, in addition to the highest mIoU achieved by HyPER-GAN, demonstrates that the proposed framework enhances photorealism without compromising the semantic integrity of the initial scenes compared to the baselines, which, while in some instances (i.e., REGEN) attempt to perform stronger updates on the rendered images (and thus achieve lower CMMD), are subject to more visual artifacts.

These observations are further illustrated in Figure \ref{fig:failures}, which compares the outputs of HyPER-GAN with the $target$ domain produced by the robust unpaired image-to-image (Im2Im) translation method (i.e., EPE). In particular, EPE often introduces artifacts, including hallucinated vegetation, geometric distortions in water surfaces, and unrealistic glossiness on vehicles, even under low-light conditions. HyPER-GAN avoids learning these artifacts and, particularly due to fewer hallucinations, leads to a higher mIoU on the PFD test set compared to EPE. In contrast, FastCUT and REGEN remain prone to unrealistic glossiness and hallucinations, as illustrated in Figure \ref{fig:baselines_comp_img}. For additional qualitative examples, see Appendix \ref{app: additional_qual}.

\begin{table}[h!]
\centering
\resizebox{1\columnwidth}{!}{%
\begin{tabular}{lccc|ccc}
\hline
\multirow{2}{*}{Model} & \multicolumn{3}{c|}{CS} & \multicolumn{3}{c}{MV} \\
 & $\downarrow$ CMMD & $\downarrow$ LPIPS & $\uparrow$ mIoU 
 & $\downarrow$ CMMD & $\downarrow$ LPIPS & $\uparrow$ mIoU \\
\hline
PFD (GTA-V)          & 5.274 & 0.000 & 48.76\% & 4.296 & 0.000 & 61.08\% \\
\hdashline
FastCUT      & 3.726 & 0.346 & 45.98\% & 3.495 & 0.255 & 58.34\% \\
REGEN        & \textbf{2.872} & 0.352 & 46.05\% & \textbf{2.386} & 0.367 & 54.90\% \\
HyPER-GAN-EO & 3.166 & \textbf{0.299} & \underline{47.77\%} & 2.992 & \underline{0.179} & \underline{59.34\%} \\
HyPER-GAN    & \underline{3.080} & \underline{0.304} & \textbf{48.06\%} & \underline{2.827} & \textbf{0.168} & \textbf{59.68\%} \\
\hline
\end{tabular}%
}
\caption{Within-engine visual realism (CMMD) and content preservation (LPIPS and mIoU) evaluation on the photorealism-enhanced images produced by FastCUT, REGEN, HyPER-GAN-EO, and HyPER-GAN on PFD (GTA-V). Bold values indicate the best, and underlined values the second-best scores. \label{tab:exp1}}
\end{table}

\begin{figure*}
	\centering
	\includegraphics[width=1\textwidth]{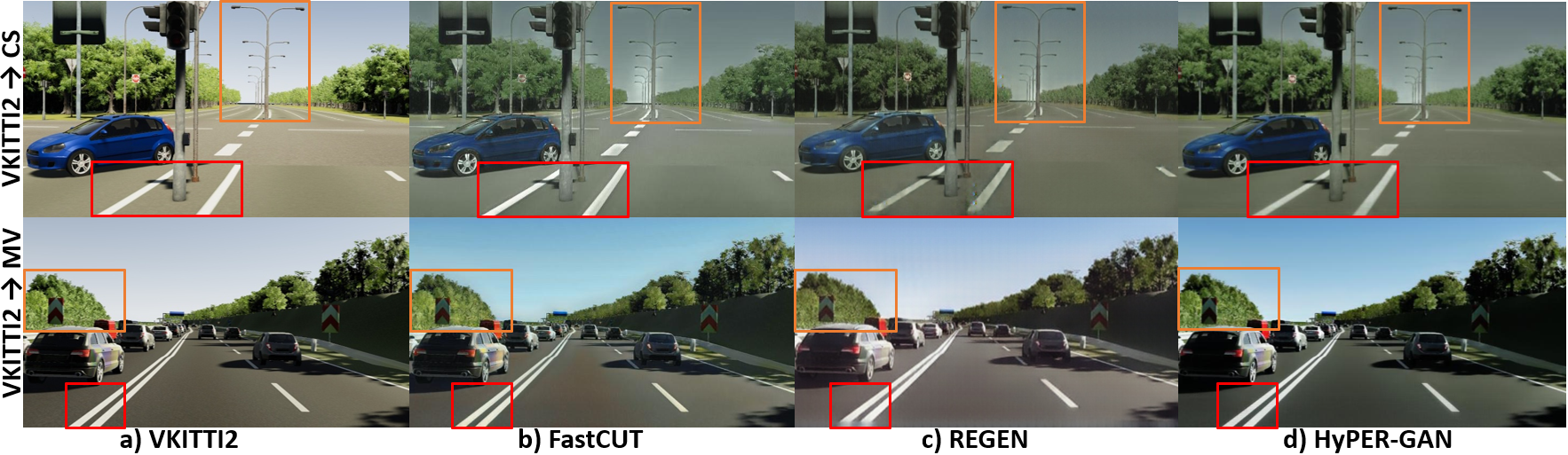}
	\caption{Translation results of a) VKITTI2 towards the real-world datasets CS and MV produced by b) FastCUT, c) REGEN, and d) HyPER-GAN.}
	\label{fig:rer_comp}
\end{figure*}

\begin{figure*}
	\centering
	\includegraphics[width=1\textwidth]{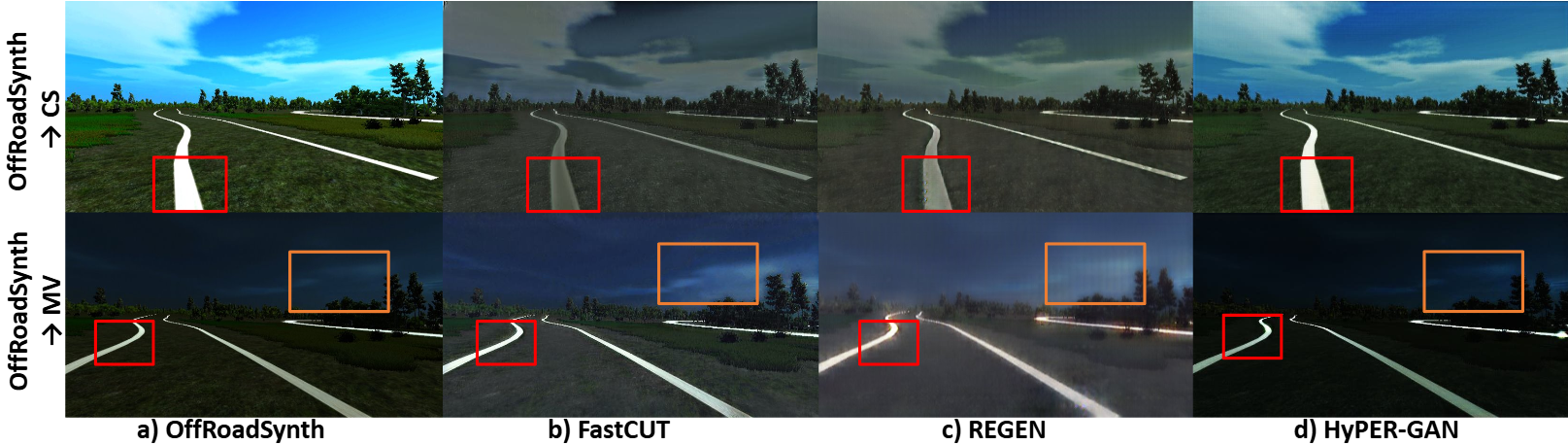}
	\caption{Translation results of a) OffRoadSynth towards the real-world datasets CS and MV produced by b) FastCUT, c) REGEN, and d) HyPER-GAN.}
	\label{fig:offroadsynth_comp}
\end{figure*}

\begin{figure}
	\centering
	\includegraphics[width=1\columnwidth]{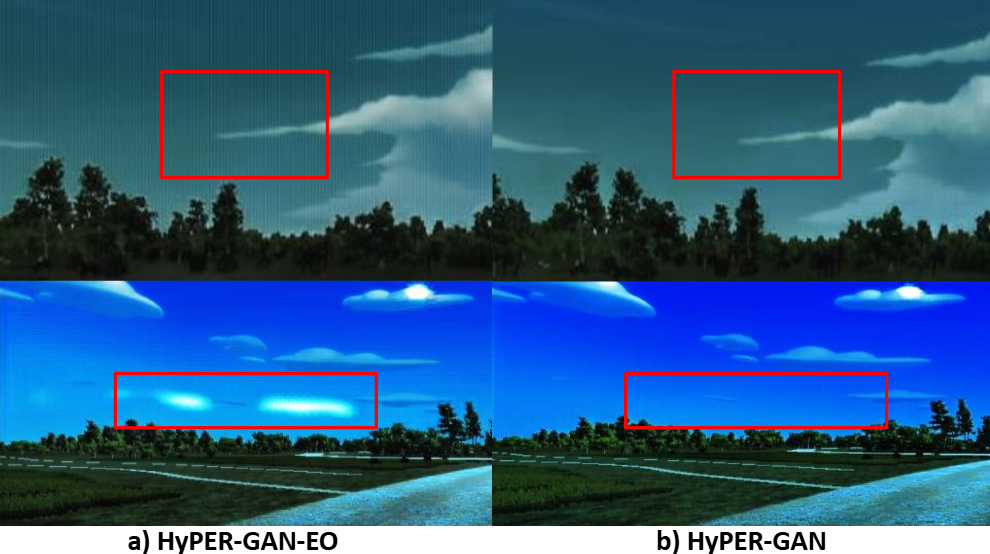}
	\caption{Translation results of a) HyPER-GAN-EO and b) HyPER-GAN on OffRoadSynth towards the real-world datasets CS (top row) and MV (bottom row).}
	\label{fig:hyperganeo_comp}
\end{figure}

\paragraph{Cross-engine Results: }  Cross-engine evaluation results on VKITTI2 and OffRoadSynth  are presented in Table \ref{tab:req_table} and Table \ref{tab:offroad_comp}, respectively. For the VKITTI2 dataset (Table \ref{tab:req_table}), HyPER-GAN again achieves the best or second-best performance across most metrics for both the CS and MV photorealism-enhanced images. In more detail, compared to FastCUT, HyPER-GAN provides superior performance across all the metrics, with the exception of the mIoU for MV. Furthermore, when compared to REGEN, HyPER-GAN leads to significantly better semantic preservation (i.e., lower LPIPS and higher mIoU). The lower (compared to HyPER-GAN) CMMD but higher LPIPS and mIoU for REGEN  suggest that while it attempts to increase visual realism more aggressively,  it generates significant visual artifacts due to the unseen (during training) environments in VKITTI2. In addition, HyPER-GAN outperforms the HyPER-GAN-EO variation across all metrics with the exception of CMMD for CS, for a similar reason as REGEN. Indeed, as illustrated in Figure \ref{fig:rer_comp}, FastCUT and REGEN have more significant visual artifacts (e.g., on the lanes of the road and the traffic poles), which justify their increased LPIPS values. Nevertheless, HyPER-GAN achieves a reduction in CMMD compared to the initial rendered VKITTI2 images, while maintaining lower LPIPS and higher mIoU scores (fewer artifacts). The tendency of the models to attempt to increase the visual realism in the unseen environments (cross-engine), similarly to the within-engine environment (PFD), and their failure in maintaining an adequate level of content preservation is further intensified for the OffRoadSynth dataset. This is because this dataset is captured in an off-road environment that significantly deviates from the urban environment depicted in the PFD dataset, which was used for training. In detail, as shown in Table \ref{tab:offroad_comp}, HyPER-GAN manages to reduce the CMMD value of the rendered images (OffRoadSynth) while maintaining the highest LPIPS and mIoU scores with instances where large margins are observed. Although FastCUT, REGEN, and HyPER-GAN-EO attempt to further reduce the CMMD, like in the within-engine experiment, they have notable issues with content preservation, leading to artifacts, distortions, and hallucinations, which are reflected at the LPIPS and mIoU metrics. For example, for FastCUT and REGEN, there is a significant impact on the mIoU, which is reduced to $67.20\%$ and $77.29\%$ respectively for the MV dataset,  compared to HyPER-GAN ($81.64\%$), indicating many visual artifacts. Indeed, as shown in Figure \ref{fig:offroadsynth_comp}, these methods are subject to a large number of artifacts in  OffRoadSynth. For the nighttime image particularly,  FastCUT and REGEN trained to translate PDF towards MV, render the image with brighter colors, which deviates from the original nighttime appearance. On the other hand, HyPER-GAN intensifies the darkness, which is the expected behavior for photorealism enhancement, since the game engine often struggles to accurately reproduce low-light conditions, such as hard shadows and nighttime illumination. Furthermore, there is a notable difference between HyPER-GAN-EO and HyPER-GAN in LPIPS for the CS dataset,   namely $0.391$ for HyPER-GAN-EO and $0.318$ for HyPER-GAN. This suggests that HyPER-GAN-EO also introduces significant artifacts. Indeed, as shown in Figure \ref{fig:hyperganeo_comp}, HyPER-GAN-EO adds several artifacts such as line patterns that appear throughout the photorealism-enhanced images compared to HyPER-GAN. Overall, the results demonstrate that HyPER-GAN achieves the best balance between visual realism and content preservation under the challenging cross-engine domain shifts. Unlike FastCUT, REGEN, and HyPER-GAN-EO, which continue to pursue larger realism improvements at the expense of introducing artifacts and semantic distortions, HyPER-GAN consistently improves visual realism while preserving the structural and semantic integrity of the initial images.

\paragraph{Temporal Consistency: } Finally, regarding the temporal consistency of HyPER-GAN on the CrowdFlow dataset, as illustrated in Table \ref{tab:optical}, the results are comparable with the ones of REGEN in \cite{regen}. Particularly, the photorealism-enhanced images lead to a slight improvement in the background due to the smoothing of small artifacts in the rendered images (i.e., anti-aliasing) and a slight increase in the error on the crowd, which is expected due to the increase in the shadows, which, in some cases, occlude some individuals of the crowd.

\begin{table}[h!]
\centering

\resizebox{1\columnwidth}{!}{%
\begin{tabular}{lccc|ccc}
\hline
\multirow{2}{*}{Model} & \multicolumn{3}{c|}{CS} & \multicolumn{3}{c}{MV} \\
 & $\downarrow$ CMMD & $\downarrow$ LPIPS & $\uparrow$ mIoU 
 & $\downarrow$ CMMD & $\downarrow$ LPIPS & $\uparrow$ mIoU \\
\hline
VKITTI2          & 4.602 & 0.000 & 46.32\% & 3.364 & 0.000 & 55.90\% \\
\hdashline
FastCUT      & 3.751 & 0.343 & 44.58\% & 3.337 & 0.282 & \textbf{53.67\%} \\
REGEN        & \textbf{3.423} & 0.446 & 41.71\% & \textbf{2.853} & 0.487 & 48.77\% \\
HyPER-GAN-EO & \underline{3.551} & \underline{0.308} & \underline{44.09\%} & 3.082 & \underline{0.252} & 50.74\% \\
HyPER-GAN    & 3.720 & \textbf{0.302} & \textbf{44.88\%} & \underline{3.079} & \textbf{0.243} & \underline{52.08\%} \\
\hline
\end{tabular}%
}
\caption{Cross-engine visual realism (CMMD) and content preservation (LPIPS and mIoU) evaluation on the photorealism-enhanced images produced by FastCUT, REGEN, HyPER-GAN-EO, and HyPER-GAN on VKITTI2. Bold values indicate the best, and underlined values the second-best scores. \label{tab:req_table}}
\end{table}

\begin{table}[h!]
\centering

\resizebox{1\columnwidth}{!}{%
\begin{tabular}{lccc|ccc}
\hline
\multirow{2}{*}{Model} & \multicolumn{3}{c|}{CS} & \multicolumn{3}{c}{MV} \\
 & $\downarrow$ CMMD & $\downarrow$ LPIPS & $\uparrow$ mIoU
 & $\downarrow$ CMMD & $\downarrow$ LPIPS & $\uparrow$ mIoU \\
\hline
OffRoadSynth & 5.586 & 0.0000 & 69.48\% & 4.475 & 0.0000 & 83.22\% \\
\hdashline
FastCUT      & \textbf{4.913} & 0.419 & 68.64\% & 4.119 & 0.363 & 77.29\% \\
REGEN        & 4.963 & 0.396 & 63.91\% & \textbf{3.912} & 0.483 & 67.86\% \\
HyPER-GAN-EO & \underline{4.914} & \underline{0.391} & \underline{68.72\%} & \underline{4.108} & \underline{0.302} & \underline{81.02\%} \\
HyPER-GAN    & 5.047 & \textbf{0.318} & \textbf{68.95\%} & 4.317 & \textbf{0.284} & \textbf{81.64\%} \\
\hline
\end{tabular}%
}
\caption{Cross-engine visual realism (CMMD) and content preservation (LPIPS and mIoU) evaluation on the photorealism-enhanced images produced by FastCUT, REGEN, HyPER-GAN-EO, and HyPER-GAN on OffRoadSynth. Bold values indicate the best, and underlined values the second-best scores. \label{tab:offroad_comp}}
\end{table}

\begin{table}[h!]
\centering
\scalebox{0.9}{
\begin{tabular}{lcc}
\hline
Model & \multicolumn{2}{c}{CrowdFlow} \\
 & Background & Crowd   \\
\hline
CrowdFlow (UE4)                    & 2.491 & 0.996 \\
\hdashline
HyPER-GAN (CS)      & 2.349 & 1.023 \\
HyPER-GAN (MV)           & 2.422 & 1.020 \\
\hline
\end{tabular}
}
\caption{Optical flow error of the initial rendered CrowdFlow dataset, the photorealism-enhanced images of HyPER-GAN (trained on PFD) towards the characteristics of CS and MV compared to the ground truth flows. The endpoint error metric is employed to calculate the error, where lower values are better. \label{tab:optical}}
\end{table}

\section{Conclusions}

In this paper, we presented HyPER-GAN, a lightweight hybrid Im2Im translation framework for real-time photorealism enhancement of rendered images. By combining a lightweight U-Net–style generator with a novel hybrid training strategy that incorporates matched real-world patches, HyPER-GAN achieves better content preservation while improving the visual realism of rendered images within the real-time constraints of the rendering pipelines of game engines. The experimental results demonstrated that the proposed framework outperforms existing lightweight paired Im2Im methods in semantic robustness and is capable of maintaining a better balance between semantic consistency and visual realism in cross-engine evaluations, making it suitable for photorealism enhancement of rendered images for CV research. Through a variation of the framework, HyPER-GAN-EO, it was also illustrated that the hybrid training approach improves semantic consistency and visual realism in within-dataset evaluation while being less prone to visual artifacts in unseen environments (cross-engine evaluation). Finally, it was demonstrated that HyPER-GAN can maintain temporal consistency. 

Future research can focus on further optimizing the real-time performance of the framework as well as improving the visual realism (i.e., CMMD) while in parallel maintaining a similar or superior level of content preservation (i.e., LPIPS and mIoU).







\appendix

\section{Real-Time Benchmarking inside UE5}\label{app: benchmarkingUE5}

In addition, we tested the performance of REGEN and HyPER-GAN within UE5, where the GPU will have to render the images of the engine as well as to infer them with the model (i.e., UE5+REGEN or UE5+HyPER-GAN). Along those lines, in Table \ref{tab:env_comparison}, the average FPS in a simple\footnote{\url{https://www.fab.com/listings/4c20896c-60bd-4090-be69-9ef4dbf1f7c2?lang=en}} (i.e., with fewer assets) and a more complex\footnote{\url{https://www.fab.com/listings/0faf8b5d-7a5f-4fee-a297-7a8efaba8896?lang=en}} environment are reported at a resolution of 1080p using a system with an Intel i7 14700F CPU, an NVIDIA RTX 4090 GPU with 24GB of VRAM, and 32GB of DDR4 system memory. From the results, it is evident that UE5+HyPER-GAN maintains an acceptable frame rate (i.e., close to $30$ FPS) in the range of $24-28$ FPS. On the other hand, UE5+REGEN results in FPS that deviate significantly from real-time performance ($30$ FPS).

\begin{table}[h]
\centering
\begin{tabular}{lcc}
\toprule
Model & Simple Env. & Complex Env. \\
\midrule
UE5        & 106.10 & 80.40 \\
\hdashline
UE5+REGEN      & 6.42   & 6.01  \\
UE5+HyPER-GAN  & 27.55  & 24.70 \\
\bottomrule
\end{tabular}
\caption{Average FPS comparison of standalone UE5 with REGEN and HyPER-GAN integrated within the engine (UE5+REGEN and UE5+HyPER-GAN, respectively) in a simple and complex environment. \label{tab:env_comparison}}
\end{table}

\section{Additional Qualitative Examples}\label{app: additional_qual}

\begin{figure*}
	\centering
	\includegraphics[width=1\textwidth]{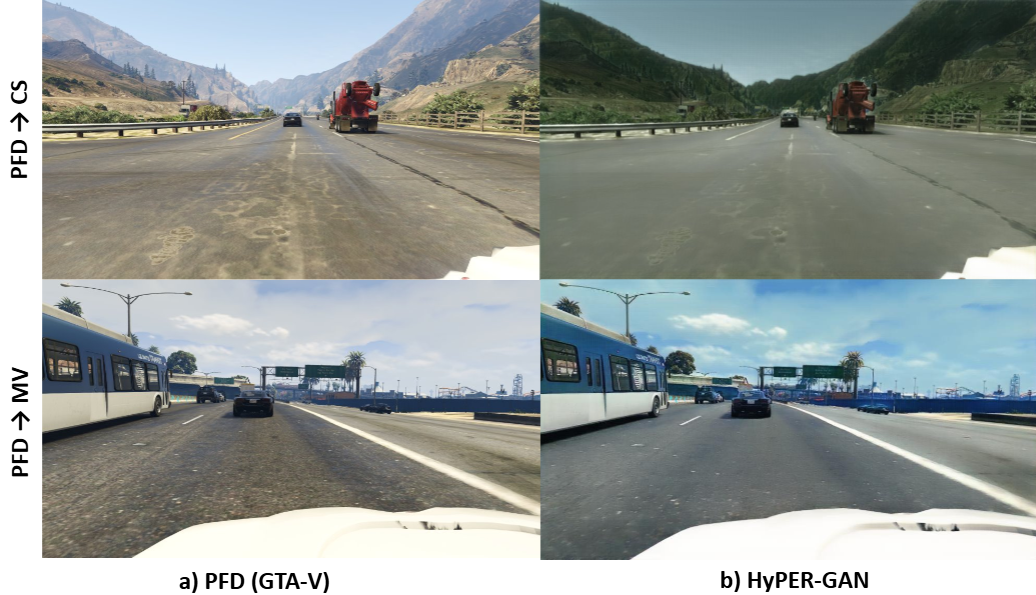}
	\caption{Translation results of b) HyPER-GAN given an input frame from the a) PFD (GTA-V) test set towards the real-world datasets CS (top row) and MV (bottom row).}
	\label{fig:additional_examples}
\end{figure*}

In Figure \ref{fig:additional_examples}, additional examples of the translation results of HyPER-GAN towards the real-world CS and MV datasets on input frames from the test set of PFD (GTA-V) are provided. Specifically, it illustrates how HyPER-GAN updates various aspects of the rendered scenes, such as mountain vegetation and road textures.

\section*{Declarations}

\textbf{Conflict of interest} The authors declare that they have no conflict of interest.



\bibliographystyle{spmpsci}      
\bibliography{bibliography}   

%
%





\end{document}